\documentclass[conference]{IEEEtran}
\IEEEoverridecommandlockouts
\usepackage{cite}
\usepackage{amsmath,amssymb,amsfonts}
\usepackage{algorithmic}
\usepackage{graphicx}
\usepackage{textcomp}
\usepackage{xcolor}
\usepackage{url}
\usepackage{booktabs}
\usepackage{tabularx}
\usepackage{array}
\usepackage{multirow}

\usepackage{comment}

\usepackage{url}
\def\BibTeX{{\rm B\kern-.05em{\sc i\kern-.025em b}\kern-.08em
    T\kern-.1667em\lower.7ex\hbox{E}\kern-.125emX}}
\begin{document}

\title{Rabies diagnosis in low-data settings: A comparative study on the impact of data augmentation and transfer learning}

\author{%
\IEEEauthorblockN{Khalil Akremi\IEEEauthorrefmark{1}, 
Mariem Handous\IEEEauthorrefmark{2},  
Zied Bouslama\IEEEauthorrefmark{3},
Farah Bassalah\IEEEauthorrefmark{2},\\
Maryem Jebali\IEEEauthorrefmark{2},
Mariem Hanachi\IEEEauthorrefmark{4},  
Ines Abdeljaoued-Tej\IEEEauthorrefmark{1}\IEEEauthorrefmark{4}}\\

\IEEEauthorblockA{\IEEEauthorrefmark{1}%
University of Carthage, Engineering School of Statistics and Information Analysis, Tunis, Tunisia. \\
Email: khalil.akremi@essai.ucar.tn}

\IEEEauthorblockA{\IEEEauthorrefmark{2}%
University of Tunis-El-Manar, Rabies Laboratory, Institut Pasteur de Tunis, Tunis, Tunisia. \\
Email: mariem.handous@pasteur.tn; ORCID: \url{http://orcid.org/0000-0002-7494-0463}}

\IEEEauthorblockA{\IEEEauthorrefmark{3}%
World Health Organization, Country Office Tunisia. \\
Email: zied.bouslama@who.int; ORCID: \url{http://orcid.org/0000-0002-9424-0704}}

\IEEEauthorblockA{\IEEEauthorrefmark{4}%
University of Tunis-El-Manar, BIMS Laboratory (LR16IPT06), Institut Pasteur de Tunis, Tunis, Tunisia.\\ Email: mariem.hanachi@gmail.com; ines.tej@essai.ucar.tn; ORCID: \url{http://orcid.org/0000-0002-1796-7897}}%
}

\IEEEpubid{%
  \makebox[\columnwidth]{%
    \parbox{\columnwidth}{%
      \scriptsize
      979-8-3315-2212-4/25/\$31.00~\copyright2026 IEEE. Personal use of this material is permitted. Permission from IEEE must be obtained for all other uses, in any current or future media, including reprinting/republishing this material for advertising or promotional purposes, creating new collective works, for resale or redistribution to servers or lists, or reuse of any copyrighted component of this work in other works.
    }%
  }%
  \hspace{\columnsep}%
  \makebox[\columnwidth]{}
}
\maketitle
\IEEEpubidadjcol

\begin{abstract}

Rabies remains a major public health concern across many African and Asian countries, where accurate diagnosis is critical for effective epidemiological surveillance. The gold standard diagnostic methods rely heavily on fluorescence microscopy, necessitating skilled laboratory personnel for the accurate interpretation of results. Such expertise is often scarce, particularly in regions with low annual sample volumes. This paper presents an automated, AI-driven diagnostic system designed to address these challenges. We developed a robust pipeline utilizing fluorescent image analysis through transfer learning with four deep learning architectures: EfficientNet-B0, EfficientNet-B2, VGG16, and Vision Transformer (ViT-B-16). Three distinct data augmentation strategies were evaluated to enhance model generalization on a dataset of 155 microscopic images (123 positive and 32 negative). Our results demonstrate that TrivialAugmentWide was the most effective augmentation technique, as it preserved critical fluorescent patterns while improving model robustness. The EfficientNet-B0 model—utilizing Geometric \& Color augmentation and selected through stratified 3-fold cross-validation—achieved optimal classification performance on cropped images. Despite constraints posed by class imbalance and a limited dataset size, this work confirms the viability of deep learning for automating rabies diagnosis. The proposed method enables fast and reliable detection with significant potential for further optimization. An online tool was deployed to facilitate practical access, establishing a framework for future medical imaging applications. This research underscores the potential of optimized deep learning models to transform rabies diagnostics and improve public health outcomes.
\end{abstract}


\section{Introduction}

Rabies remains an endemic and fatal zoonotic disease, responsible for over 59,000 human deaths annually, with the majority of cases occurring in Africa and Asia \cite{2}. In Tunisia, sustained efforts have been deployed to control and prevent the spread of rabies \cite{bouslama2021molecular}. However, since 2022, the country has experienced a marked increase in reported cases, bringing renewed attention to the disease as a critical public health concern. This resurgence has also intensified media coverage and sparked growing interest within the scientific community.

Among the key components of an effective rabies control program, surveillance plays a pivotal role and relies heavily on robust laboratory capacities. Rabies is a pleiotropic disease, and its confirmatory diagnosis requires laboratory-based methods \cite{rab}. The gold standard for diagnosis is the Fluorescent Antibody Test (FAT), which detects the rabies virus antigen, specifically the nucleocapsid protein. This technique involves preparing brain tissue smears on glass slides, fixing them with acetone, and staining them using an anti-rabies nucleocapsid antibody conjugated with a fluorescein fluorochrome. Accurate reading and interpretation of the stained slides require highly trained and experienced laboratory personnel \cite{3}.

The existing rabies diagnostic process encounters several major obstacles. Post-mortem tissue samples often suffer from degradation and inconsistent preservation quality, leading to unreliable test results. The condition of samples varies significantly depending on storage conditions, transportation duration, and handling procedures. Microscopic analysis is frequently complicated by the presence of salt crystals and auto-fluorescence phenomena, which can mask true positive signals or create false positives. These artifacts significantly complicate the interpretation process and require expert knowledge to distinguish them from viral antigens. Furthermore, rural regions face a critical shortage of trained experts capable of performing accurate microscopic analysis. Rapid diagnosis is essential to saving lives \cite{asma2024child}.

We utilized laboratory-collected data to support reliable rabies diagnosis. To achieve this, we leveraged digitized biological data and proposed a robust, automated solution. The implementation of such a system aims to improve the current situation by reducing diagnostic turnaround time. Furthermore, it addresses the expert shortage by providing reliable diagnostic capabilities in regions where specialized personnel are not readily available, ensuring consistent and accurate diagnoses across all geographical areas.

Our primary objective is to accurately classify rabies-infected cells through automated image analysis. We employed a comprehensive methodology initiating with image annotation using fine-tuned YOLO models, facilitated by the Roboflow platform. Subsequently, we investigated advanced deep learning techniques through transfer learning, utilizing pre-trained models such as EfficientNet-B0 and B2, VGG-16, and Vision Transformer (ViT). To enhance model robustness and prevent overfitting, data augmentation techniques were applied throughout the training process. Finally, we deployed an online solution accessible at \footnote{\url{http://huggingface.co/spaces/huggingkhalil/efficientnet-classifier}}.

The interpretation of FAT results can be challenging, particularly when dealing with decomposed samples or species prone to diagnostic artifacts. The integration of AI-based tools offers promising support by enabling rapid and standardized diagnosis, reducing subjectivity, and minimizing human reading errors. This is especially beneficial for laboratories processing large volumes of samples, where prolonged microscope usage can lead to eye strain and fatigue. This necessity was underscored in Tunisia recently, where a significant surge in sample submissions occurred over a short period, highlighting the urgent need for automated assistance in diagnostic workflows.

\section{Background}

The current rabies diagnostic process faces several critical challenges. One major obstacle is the variability in sample quality, which can significantly affect diagnostic reliability. Accurate interpretation requires highly trained personnel with extensive expertise, as the reference diagnostic techniques for rabies are still performed manually. Introducing an automated approach could represent a pivotal step toward modernizing rabies diagnostics, making the process more efficient, accessible, and rapid.

Thresholding algorithms encompass a range of methods tailored for different image characteristics \cite{otsu1975threshold,otsu1979threshold,zack1977automatic,huang1995image,ridler1978picture,li1993minimum,kapur1985new,hero2002alpha}. While established tools exist for evaluating thresholding methods—such as open-source platforms like ImageJ/Fiji \cite{abramov}, QuPath \cite{bankhead2017qupath}, and CellProfiler \cite{mcq}, or proprietary software like MetaMorph and NIS-Elements—they are poorly suited to our immediate need for a fully automated, rapid diagnostic aid. These solutions often require significant manual intervention, expert parameter tuning, or are not designed for seamless integration into a clinical workflow. Consequently, our focus shifts towards an end-to-end deep learning approach that can directly learn optimal feature representations from the data itself, overcoming the limitations of these manual and semi-automated techniques.

\section{Data and methods}

\subsection{Dataset collection and description}

Images were obtained from wells of microplates used in the Fluorescent Antibody Virus Neutralisation (FAVN) test, performed at the Rabies Laboratory of the Pasteur Institute of Tunis, following the recommendations of WOAH \cite{x}, WHO \cite{xx}, and Cliquet et al. \cite{cliquet}. Staining was performed using the Fujirebio anti-Rabies Nucleocapsid Conjugate. A Leica DMIL LED fluorescence microscope equipped with a 10× objective was used for image acquisition. Images were saved in PNG format. 

We collected 155 microscopic images in February 2025, comprising 123 positive samples (rabies-infected) and 32 negative samples. This imbalanced distribution (79.4\% positive vs. 20.6\% negative) required careful handling through stratified sampling and weighted loss functions. The dataset was partitioned using stratified splitting to preserve class proportions across training, validation, and test sets: Training set (70\%); Validation set (15\%); Test set (15\%). All experimental code and data processing scripts are available on GitHub\footnote{\url{https://github.com/khalil-akremi/rabies-classification}}.

\subsection{Preprocessing using YOLO and Roboflow}

\begin{figure}[htbp]
    \centering
    \includegraphics[width=0.23\textwidth]{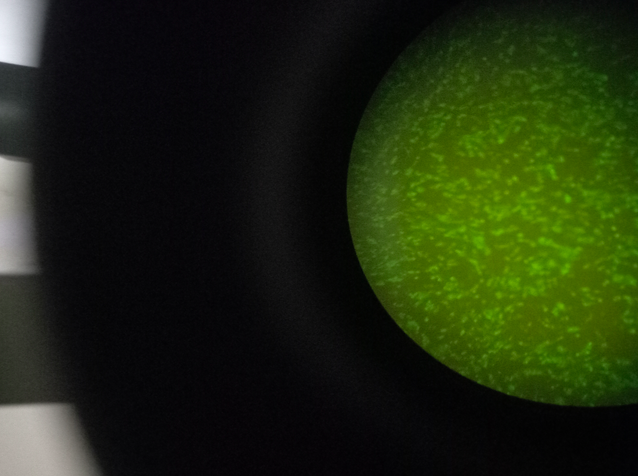} 
    \includegraphics[width=0.24\textwidth]{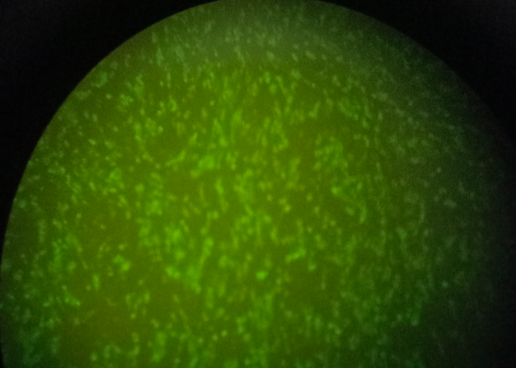} 
    \caption{Comparison of original (left) and YOLO-annotated (right) microscopic images}
    \label{fig:crop}
\end{figure}

Given the limitations of conventional detection methods, we adopted a robust strategy combining manual annotation with modern object detection techniques. We employed the Roboflow platform for precise image annotation, delineating detailed bounding boxes around regions of interest within the microscopic images \cite{alex}. Subsequently, we fine-tuned the YOLOv8 model \cite{yolo} to automatically localize and extract relevant regions, as illustrated in Fig. \ref{fig:crop}. This preprocessing pipeline yielded two parallel datasets: raw (uncropped) images representing the full-field images (FFI); cropped (YOLO-extracted) regions of interest called segmented diagnostic patches (SDP). The two datasets were subjected to identical augmentation and training protocols to facilitate comparative analysis. 

\subsection{Data augmentation strategies}

To mitigate the limitations of dataset size and severe class imbalance, extensive data augmentation was applied exclusively to the training set. Three augmented instances were generated for each original training image, resulting in a fourfold ($4\times$) dataset expansion. This augmentation process resulted in an identical class distribution for both the {FFI} and {SDP} datasets. Specifically, the positive class was expanded from $86$ original images to a total of $344$ (incorporating $258$ augmented variations), while the negative class increased from $22$ original images to $88$ (including $66$ augmented variations). Consequently, each dataset variant comprised a total of $432$ training images. The validation and test sets remained unaltered, to ensure unbiased evaluation. For the training phase, three specialized augmentation strategies were implemented: 

\begin{itemize}
    \item {TrivialAugmentWide} applies random transformations from a predefined set, introducing necessary variation while maintaining image integrity—a critical factor for preserving subtle fluorescence patterns \cite{muller}.
    \item {Geometric \& Color Transformations} incorporates random rotation ($30^{\circ}$) and color jitter to simulate variations in specimen orientation, staining intensity, and laboratory lighting conditions—common challenges in fluorescence microscopy \cite{shorten}.
    \item {Spatial \& Blur Transformations} employs horizontal flipping and Gaussian blur to account for spatial diversity and focus inconsistencies encountered during the microscopic examination of rabies samples \cite{blur1}.
\end{itemize}

\subsection{Transfer learning architectures}

We evaluated four representative architectures:

\begin{itemize}
    \item {VGG16:} A classic architecture with proven efficacy in medical imaging, characterized by its substantial parameter count (138M) and the use of uniform $3\times3$ convolutional filters followed by fully connected layers  \cite{tou}.
    \item {EfficientNet-B0 \& B2:} Known for achieving an optimal balance between accuracy and computational efficiency. We selected two scaled variants (B0: 5.3M parameters; B2: 8.8M parameters) that share a unified architecture while differing in capacity. The EfficientNet family \cite{effi} utilizes compound scaling to systematically optimize network depth, width, and resolution, employing inverted residual blocks as efficient building components. This design delivers state-of-the-art performance with significantly fewer parameters and FLOPs compared to conventional networks like VGG16 \cite{agg}.
    \item {Vision Transformer (ViT-B-16):} A modern attention-based architecture for image classification (86.6M parameters) that relies on patch-based multi-head self-attention mechanisms rather than traditional convolutional operations.
\end{itemize}

These models represent distinct evolutionary paths in deep learning: depth efficiency (EfficientNet), spatial hierarchy (VGG), and attention mechanisms (ViT), providing diverse architectural perspectives for our comparative analysis of medical image classification performance.

\subsection{Training configuration and hyperparameters}

All models were pretrained on ImageNet and fine-tuned for binary rabies classification. The implementation involved freezing the pretrained backbone weights and training only the classification head on our specific dataset.\footnote{The hyperparameter configuration adopted for the training process relies on the AdamW optimizer, initialized with a specific learning rate of 1e-4 and a weight decay of equivalent magnitude to ensure effective regularization, while momentum parameters are maintained at their standard values.} To manage the optimization trajectory, the training phase is bounded between 25 and 30 epochs.
The batch size is generally set to 32, with a reduction to 16 for the Vision Transformer architecture to accommodate GPU memory constraints. Regarding data processing, all input images are resized to 224x224 pixels in RGB format and normalized according to standard ImageNet statistics. To address class imbalance issues, the model minimizes a weighted Cross-Entropy loss function.\footnote{We set negatif class weight to 2.4545 in order to compensates minority class; positif class weight is set to 0.6279 which is proportional to class frequency}.

Furthermore, to ensure statistical robustness and reproducibility, the evaluation employs a stratified $K$-Fold cross-validation method with $K=3$. Indeed, given the class imbalance, the adoption of stratified $K$-fold cross-validation was imperative to ensure that each fold preserved the original class distribution ($79\%$ positive and $21\%$ negative). By preventing the formation of folds containing exclusively positive or negative samples and guaranteeing consistent minority class representation across all splits, this strategy enables robust model evaluation despite the scarcity of negative samples. 

To further mitigate the impact of class imbalance during training, class weights were computed inversely proportional to class frequencies.
By amplifying the loss contribution of rare samples, this strategy counteracts the optimization bias toward the majority class, ensuring meaningful feature learning for both categories.

\subsection{Cross-validation and model selection}

Stratified 3-fold cross-validation was employed to obtain robust performance estimates across all 12 model-augmentation-dataset configurations (4 architectures $\times$ 3 augmentation strategies on 2 dataset types: FFI/SDP). Performance metrics (accuracy, AUC, precision, recall, F1-score) were averaged across folds to identify the optimal configuration. 

Following cross-validation analysis, 
the optimal configuration was retrained on the complete augmented training set (432 images, all 3 folds combined) to maximize learning capacity prior to final evaluation on the held-out test set.

\section{Results}

\begin{figure}[!ht]
\centering
\includegraphics[width=0.5\textwidth]{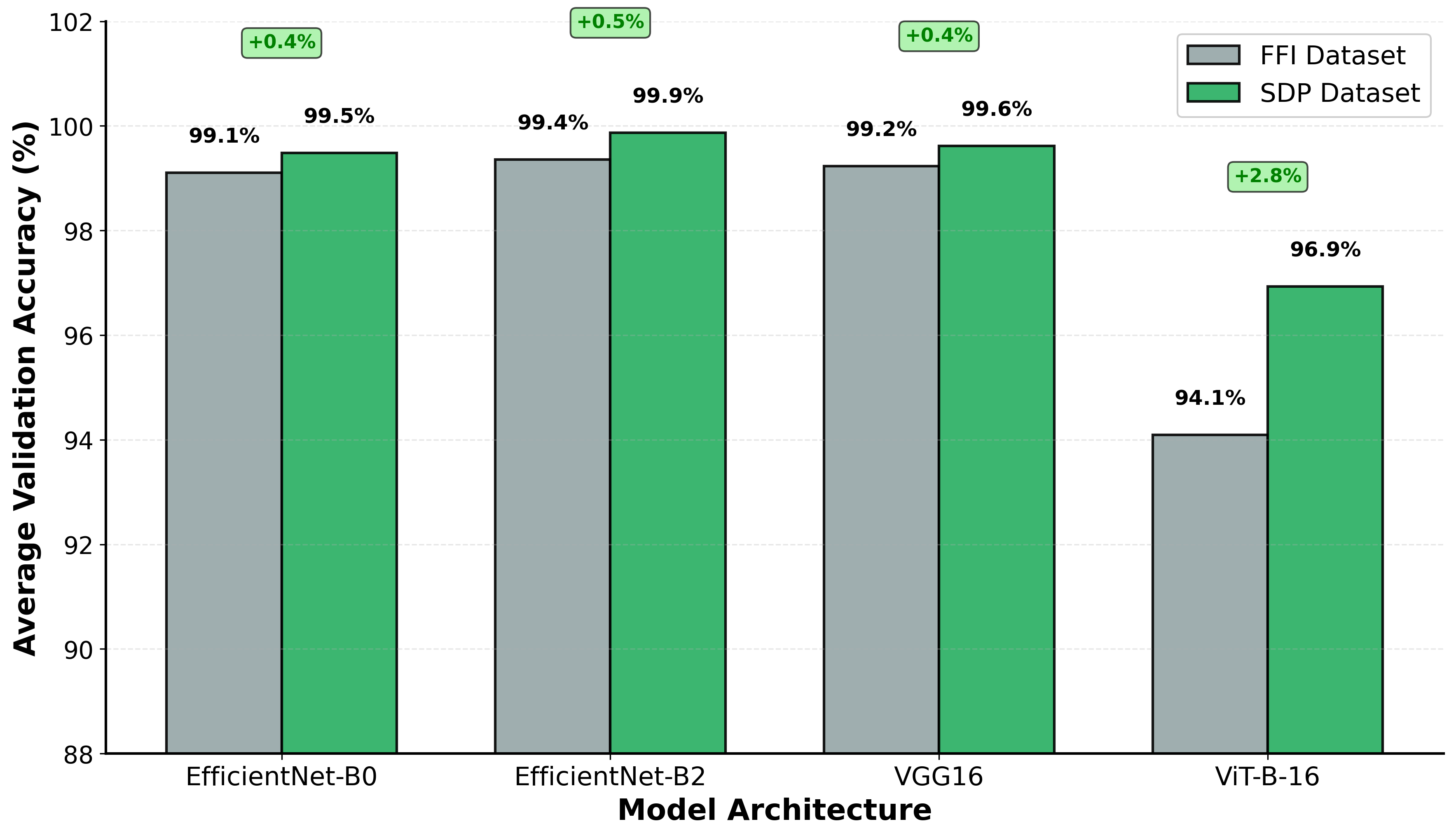}
\caption{Impact of image preprocessing on model performance}
\label{fig:model_performance}
\end{figure}

We applied modern artificial intelligence techniques to address critical public health challenges in rabies diagnosis. Through systematic evaluation of four deep learning architectures (EfficientNet-B0, EfficientNet-B2, VGG16, ViT-B-16) across three augmentation strategies and two dataset variants (FFI vs. SDP), we identified EfficientNet-B0 (SDP + Geometric \& Color Transformations) as the optimal configuration based on cross-validation performance. Indeed, Fig. \ref{fig:model_performance} summarizes model performance across all configurations. EfficientNet-B2 with TrivialAugmentWide achieved perfect classification performance, demonstrating exceptional consistency and balanced metrics. However, considering computational efficiency and generalization stability across multiple augmentation strategies, EfficientNet-B0 (SDP + Geometric \& Color Transformations) was selected for final deployment (Fig. \ref{fig22} and Fig. \ref{fig2} describe the models performance analysis across augmentation strategies).\\

\begin{figure*}[!ht]
\centering
\includegraphics[width=\textwidth]{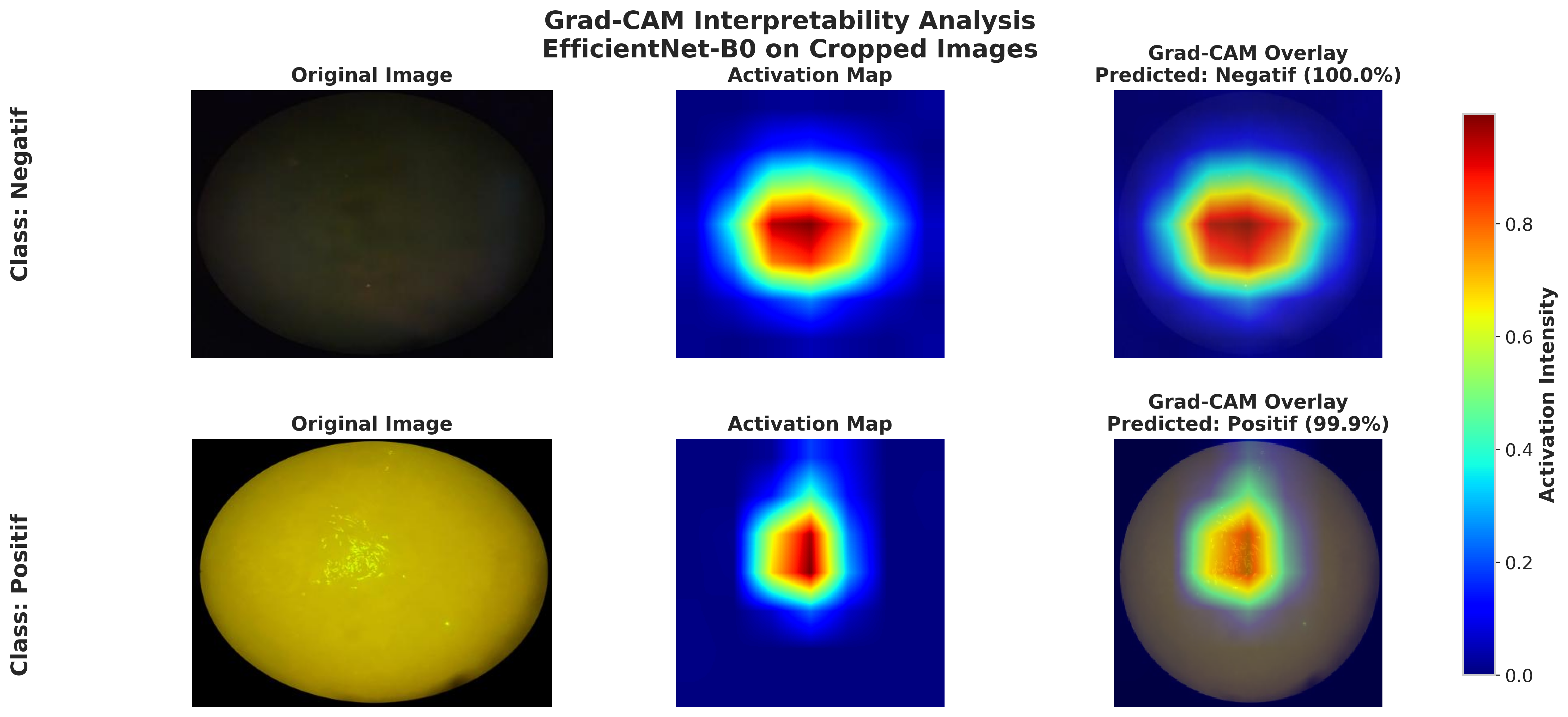}
\caption{Grad-CAM Model Interpretability: Visualization of learned features highlighting rabies-infected regions}
\label{fig:gradcam}
\end{figure*}

\begin{figure*}[!ht]
    \centering
    \includegraphics[width=0.8\textwidth]{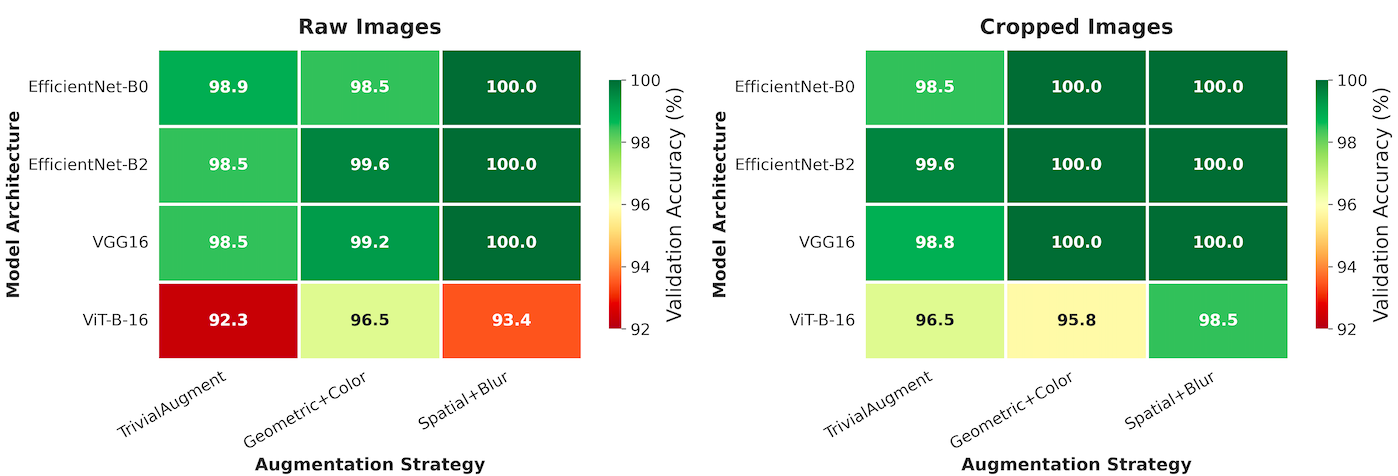}
    \caption{Model performance across all aonfigurations: Accuracy evolution during training}
\label{fig22}
\end{figure*}

\begin{figure*}[!ht]
\centering
    \includegraphics[width=0.8\textwidth]{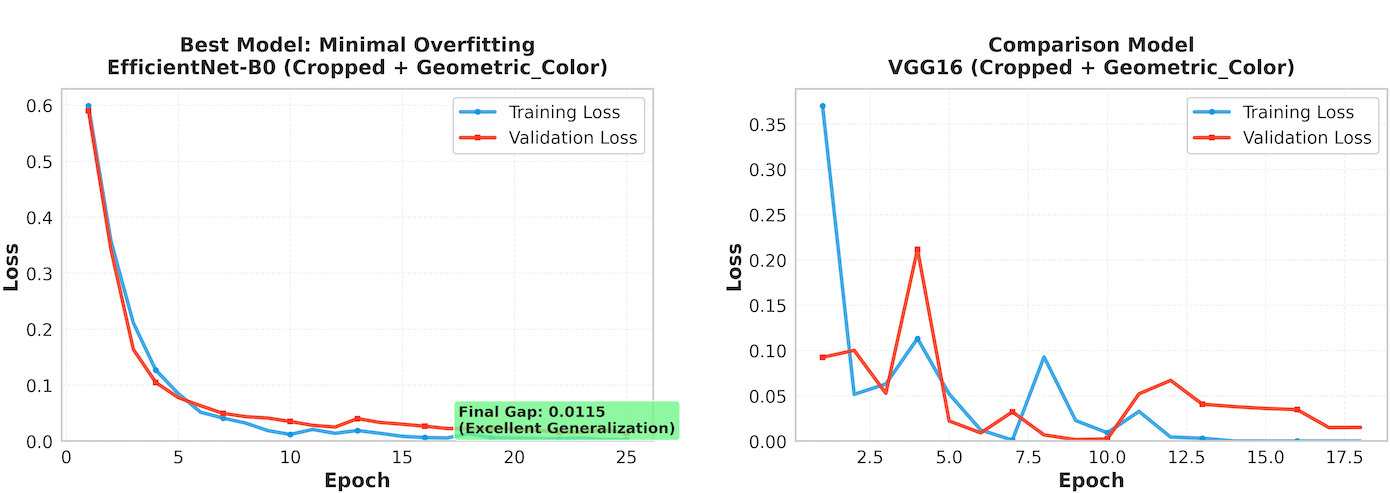}
    \caption{Training and validation Loss curves: Overfitting detection across model architectures and augmentation strategies}
    \label{fig2}
\end{figure*}

Gradient-weighted class activation mapping (Grad-CAM) visualizations were generated for representative samples from both classes, providing qualitative evidence that the model learned task-relevant features rather than memorizing dataset artifacts, see Fig. \ref{fig:gradcam}. 
The final deployment of our rabies detection system leverages web-based technologies to ensure accessibility and usability for medical professionals. We implemented the solution using the Gradio framework for the user interface and deployed it on Hugging Face Spaces for reliable cloud hosting.

\section{Discussion}

The FAVN technique was employed for image acquisition rather than the standard FAT, as positive samples derived from FAVN yield clearer, higher-quality imagery, rendering them more suitable for the initial phase of machine learning model training. However, this superior image quality was constrained by a limited dataset size ($155$ images) and severe class imbalance ($79.4\%$ positive). To mitigate these limitations and prevent overfitting, the implementation of robust data augmentation and stratified cross-validation proved essential.\\

TrivialAugmentWide proved critical for achieving optimal performance, effectively enhancing data diversity while preserving essential fluorescence patterns and minimizing the introduction of destructive noise. Alternative models exhibited significant limitations: $VGG16$ displayed inconsistent generalization ($100\%$ accuracy on two augmentation sets but only $75\%$ on Geometric \& Color Transformations). Furthermore, $ViT-B-16$ consistently underperformed ($75\%$ accuracy across all conditions) and demonstrated unstable convergence, likely attributable to the insufficiency of training data required for attention-based architectures.\\

Each model underwent fine-tuning using our annotated rabies dataset under various augmentation strategies to identify the optimal deployment configuration. Despite achieving $100\%$ accuracy and minimized loss values in specific configurations, the available dataset presented significant challenges. The dataset size was insufficient for effectively training deep learning classification models from scratch, necessitating the use of transfer learning with $ImageNet$-pretrained weights. Additionally, variability in image resolution and staining quality in certain samples limited the models' ability to extract the fine-grained features necessary for accurate rabies classification. The acquisition of a larger, high-resolution dataset would significantly improve these results and enhance clinical viability.\\

A major challenge in accurate rabies diagnosis using the Fluorescent Antibody Test (FAT) is the requirement for highly trained technical personnel capable of distinguishing subtle differences between positive and negative samples under a fluorescence microscope. This task demands not only technical proficiency but also extensive experience and precision—standards that are difficult to maintain consistently across laboratories, particularly in resource-limited settings. Artificial Intelligence (AI) offers a robust solution by assisting or potentially automating the interpretation of diagnostic results. Leveraging computer vision and deep learning, AI models can be trained on extensive datasets of labeled samples to discriminate between rabies-positive and negative patterns with high accuracy. This approach not only reduces dependence on specialized personnel but also ensures faster, more consistent, and potentially more objective diagnostic outcomes. Consequently, AI can facilitate expanded access to reliable rabies testing, particularly in regions facing a scarcity of trained experts.

\section{Conclusion}

This work addressed the challenge of creating a decision-support tool despite limitations in dataset size ($155$ images), severe class imbalance ($79.4\%$ positive), and quality—working with a constrained collection of microscopy images exhibiting variable resolution and staining conditions. Through systematic evaluation of three data augmentation strategies cross-validated with four deep learning models (EfficientNet-B0/B2, VGG16, ViT-B-16) via transfer learning, we established an optimal configuration. The EfficientNet-B0 model (SDP + Geometric \& Color Transformations) achieved robust classification performance while maintaining computational efficiency and stable generalization capabilities despite data constraints. \\

The implemented solution provides critical decision support for medical professionals, automating classification tasks that previously required specialized expertise. Its cloud-based deployment via Gradio and Hugging Face Spaces ensures practical accessibility, and the system is now actively used by our diagnostic teams in Tunisia. The development of an online, open-access tool for determining the infection status of diagnostic images offers an accessible and scalable solution for laboratories worldwide to support and confirm rabies diagnoses. \\

This study highlights the potential of AI to provide tangible solutions for critical public health challenges in resource-limited settings. Future work will expand datasets through multi-institutional collaborations, incorporate additional image variants (FAT technique, multiple staining protocols), enhance real-time diagnostic workflows, and validate clinical utility through prospective studies to further support clinical decision-making.

\section*{Acknowledgment}

The authors gratefully acknowledge Dr.~Sonia Kechaou, Dr.~Emna Harigua, Dr.~Hela Sellami, and Prof.~Alia Benkahla for their valuable advice. They also thank Dr.~Habib Kharmachi for his support, and Mr.~Omar Jlassi, Mrs.~Jihen Bensalem, Mrs.~Chaima Nouioui, Mrs.~Oumayma Roueg, Mrs.~Rihab Khlif, Mr.~Mohamed Soltani, and Mr.~Khaled Ghouili for their technical assistance. The authors thank the anonymous reviewers for their insightful feedback and constructive suggestions. 

\section*{Author Contributions}

Conceptualization: IA-T, MHand, KA, ZB, MHana; methodology and design: IA-T, MHand, KA; data collection: MHand, FB, MJ, KA, IAT; investigation: MHand, MHana, ZB, KA, IA-T; formal analysis and interpretation of results: IA-T, KA; draft manuscript preparation: MHand, IA-T, KA; revised the manuscript: ZB, MHana; supervision: IA-T. All authors reviewed the results and approved the final version of the manuscript.

\end{document}